\documentclass{paper}
\usepackage{amsmath,amsthm,mleftright}
\usepackage{amssymb,mathtools}
\DeclareMathAlphabet\mathcal{OMS}{cmsy}{m}{n}
\DeclareMathAlphabet\mathbfcal{OMS}{cmsy}{b}{n}
\usepackage{tikz-cd}
\usepackage{xspace}
\usepackage{makecell}
\usepackage{graphicx,booktabs}
\usepackage{ragged2e}
\usepackage{booktabs,multirow,tabularx}
\usepackage{paralist}
\usepackage{hyperref}
\hypersetup{%
    pdfmenubar=true,       
    pdffitwindow=false,     
    pdfstartview={FitH},    
    pdftitle={Image Reconstruction for Multispectral CT},
    colorlinks=true,       
    linkcolor=red,          
    citecolor=red,        
    filecolor=magenta,      
    urlcolor=cyan,           
    pdfborder = {0,0,0}
}
\usepackage{acro}
\usepackage{cleveref}
\usepackage{algorithm}
\usepackage[noend]{algpseudocode}
\makeatletter
\def\BState{\State\hskip-\ALG@thistlm}
\makeatother
\Crefname{equation}{}{}
\crefname{equation}{}{}
\usepackage[title]{appendix}
\usepackage{todonotes}
\usepackage{inputenc}
\theoremstyle{plain}

\theoremstyle{definition}

\theoremstyle{remark}

\usepackage{bm}
\usepackage{adjustbox}

\begin{document}

\title{GUMP-Net: An interpretable model-data-driven intelligent algorithm for multi-class pelvic segmentation}
\author{Liheng Wang\thanks{State Key Laboratory of Mathematical Sciences, Academy of Mathematics and Systems Science, Chinese Academy of Sciences, Beijing 100190, China; University of Chinese Academy of Sciences, Beijing 100190, China.},  
Yinghui Zhang\thanks{State Key Laboratory of Mathematical Sciences, Academy of Mathematics and Systems Science, Chinese Academy of Sciences, Beijing 100190, China.}, 
Licheng Zhang\thanks{Department of Orthopedics, The Fourth Medical Center of Chinese PLA General Hospital, Beijing 100048, China; National Clinical Research Center for Orthopedics, Sports Medicine and Rehabilitation, Beijing 100048, China.},  \\
Hailin Xu\thanks{Department of Trauma and Orthopedics, People's Hospital Peking University, Beijing 100044, China.}, 
Qiyong Cao\thanks{Department of Orthopedics and Traumatology, Beijing Jishuitan Hospital, Capital Medical University, Beijing 100035, China}, 
Chong Chen\footnotemark[2]
}

\maketitle

\begin{abstract}
Pelvic segmentation is one of the most important and fundamental research problems in precise and intelligent diagnosis and treatment, as well as surgical planning and navigation for pelvic fractures. By combining an improved geodesic active contour model with deep neural networks, we propose GUMP-Net, an interpretable model-data-driven intelligent algorithm for multi-class pelvic segmentation, in which three network modules are designed to constitute the overall segmentation framework together: the object detection module for automatic level set initialization, the edge detector module for learning an anatomy-aware edge detector function and the iteration module for deep level set evolution. Leveraging the advantages of level set representation and deep learning, GUMP-Net shows more accurate, robust and consistent segmentation performance, especially in small training data situation, compared to the state-of-the-art methods. Extensive experiments on pelvic datasets demonstrate the rationality and effectiveness of the proposed algorithm. Further experiments extended to ankle dataset indicate broader applications to other anatomies. The proposed algorithm not only provides an efficient segmentation method for complex fracture reduction, but also gives an interpretable geometric perspective for understanding deep learning segmentation.
\end{abstract}


\section{Introduction}
\label{sec:introduction}
Bone segmentation is a critical procedure in orthopedic medicine. For instance, the pelvis, which mainly consists of the \textit{left hip}, \textit{right hip} and \textit{sacrum}, maintains the balance of the body, protects the abdominal organs/nerves and constitutes an important supporting structure. Some pathologies such as pelvic fractures, in the most severe situations, could be very fatal for the patients. Therefore, accurate and efficient segmentation is of significant importance to trauma diagnosis, preoperative planning, surgical reduction and postoperative evaluation \cite{han2021fracture}. In the past decade, the success of deep learning methods (DLMs) has led a trend to apply this paradigm to diverse medical areas, including the pelvic segmentation \cite{Liu2020,Liu2023}. While the researchers are pursuing higher accuracy and efficiency, some clinicians are also calling for interpretability in the decision process of deep learning, so as to enhance human comprehension in clinical practice \cite{Tikhomirov2024,Toruner2024}.

Among numerous segmentation methods, active contour models (ACMs), or sometimes also formulated as level set models, are one of the most widely used approaches for their mathematical interpretability. Typically, ACMs optimize a customized energy functional that incorporates image features and geometric priors, such as the contour length, curvature and region area. These models offer an intuitive segmentation process that the contour gradually evolves towards the targets to obtain final results, which is relatively transparent, explainable and easy to accept. 

Some ACMs employ the image \textit{intensity} or \textit{gradient norm} to construct the energy functional, such as Chan--Vese (CV) model \cite{Chan2001} and geodesic active contour (GAC) model \cite{Caselles1995}. The CV model approximates the image in a piecewise-constant manner and is more suitable for homogeneous images, followed by many improvements for inhomogeneous images, such as local intensity extraction \cite{Li2008,MIN201969}, piecewise polynomial approximation \cite{Chen2017} and Jeffreys divergence based model \cite{Han2020,LIU2026112384}. The GAC model minimizes the contour length weighted by an edge detector function (EDF), which guides the active contour to find target boundaries with large gradient norm. Furthermore, some studies focused on modifying GAC model to better accommodate bone segmentation, by adding global shape constraints \cite{Jia2006} or incorporating additional orthopedic domain knowledge \cite{WANG2026113049}. Despite their good interpretability, the ACMs still suffer from the sensitivity to the image quality, initial conditions and hyper-parameters, and lag behind DLMs in utilizing massive amounts of data. 

The DLMs for image segmentation, from convolutional neural network (CNN) based to Transformer based architectures, have advanced drastically over the past decade to meet diverse medical demands. Representative CNN-based models like DeepLabv3+ \cite{Chen2018} and U-Net \cite{Ronneberger2015} have set performance benchmarks for biomedical image segmentation. Numerous variants of U-Net, including Attention U-Net \cite{Oktay2018}, nnU-Net \cite{Isensee2021} and so on \cite{Zhou2020,Li2018}, have been proposed to enhance the capability from different aspects. Meanwhile, Transformers \cite{Vaswani2017,Liu2021} have been adapted to medical image segmentation for the ability to capture long-range dependencies, such as Swin-Unet \cite{Hu2021}. After that, foundation models such as SAM \cite{SAM} designed a segmentation task with interactive prompts such as points and boxes. MedSAM \cite{MEDSAM} and \cite{SAB} further fine-tuned SAM on different medical image datasets. In contrast with ACMs, DLMs are rather black boxes which approximate the segmentation mapping by big data and optimization, lacking explicit geometric intuitions. 

To address this, a growing body of research \cite{HoangNganLe2020,Gui2023} has explored to integrate ACMs and DLMs, aiming to simultaneously achieve high performance and improve model interpretability. These integrations can be broadly categorized into three types. The first type reformulates the ACM as post-processing steps, where the initial contour \cite{Tian2023,Yang2024} and/or hyper-parameters \cite{Hatamizadeh19} used in the ACM are predicted by a \textit{pre-trained} network and are then refined by the ACM to get a final segmentation. However, the final results are still inevitably limited by the disadvantages of the ACM. The second type reformulates the energy functionals as loss functions \cite{Ma2021b,Kim2020}, which implicitly embeds geometric priors into the network parameters during training, but still remains black-box segmentation. The third type, sometimes called \textit{algorithm unrolling} \cite{Monga2021}, reformulates iterative algorithms as network blocks. Some examples include the full approximation scheme (FAS) U-Net \cite{Zhang2022} and the PottsMGNet \cite{Tai2024}. In addition, several articles represented the activation functions as optimization problems, such as the Softmax and the Rectified Linear Unit (ReLU), and then added extra penalty terms to regularize the network behaviors \cite{Liu2022,Jin2024}. 
Recent literature has combined DLMs and ACMs in an end-to-end fashion \cite{Zhang2020,Hatamizadeh2020}, where the initial contour and parameters in those ACMs are still predicted by networks, but distinct from the first type, the corresponding contour iterations are included in the training process. It is also worth noting that a class of methods \cite{Peng2020DeepSF,Yang2023BoxSnakePI} mimic the snake model by predicting a fixed number of points on the next contour, which might be limited by the unchangeable contour topology and struggle to find the inner holes of an object. In general, the unrolling paradigm effectively embeds mathematical structural priors into the DLMs while maintaining data-driven ability. Recently, ML‑DSVM+ \cite{HAN2023109076} proposed a meta‑learning based deep SVM+ framework to improve single‑modal computer‑aided diagnosis via bidirectional knowledge transfer. MTD‑Net \cite{ZHANG2026112697} introduced a multi‑task discriminative network with uncertainty estimation for choroidal neovascularization segmentation.

\textit{\textbf{Contributions.}} We propose an interpretable model-data-driven intelligent algorithm for multi-class pelvic segmentation named GUMP-Net, which includes three network modules with interpretable sub-tasks: the object detection module (ODM) for automatic level set initialization, the edge detector module (EDM) for learning an anatomy-aware edge detector function and the iteration module (IM) for deep level set evolution, addressing the black-box segmentation problem to some extent. 

Specifically, we reformulate the iteration steps of an improved GAC model into the trainable IM to evolve the multi-channel level set function (LSF), where each channel is assigned to predict the LSF of a certain anatomy. After the iteration, the zero-level set of the final LSF yields the corresponding segmentation boundary. Instead of the traditional edge detector function (EDF), we design the novel EDM to predict a learned EDF. As for the initial LSFs, we adopt the ODM to predict the bounding boxes and then construct the corresponding initial LSFs. 

The design of the novel EDM incorporates additional structural information from an improved EDF, which helps the learned EDF suppress unconcerned responses and highlight target features, effectively guiding the contour evolution.

GUMP-Net outperforms the compared methods, including CNN-based benchmarks (DeepLabv3+, U-Net, Attention U-Net, nnU-Net), Transformer-based Swin-Unet and algorithm unrolling methods (FAS-UNet, PottsMGNet) in the task of multi-class pelvic segmentation, especially exhibiting better stability and generalization ability in the situation of small training data. GUMP-Net also achieves better visual performance compared with foundation models such as SAM and MedSAM. GUMP-Net also achieves good performance in the multi-class ankle segmentation, which suggests broader applications in other anatomies.

\begin{figure}[htbp]
\includegraphics[width=0.99\textwidth]{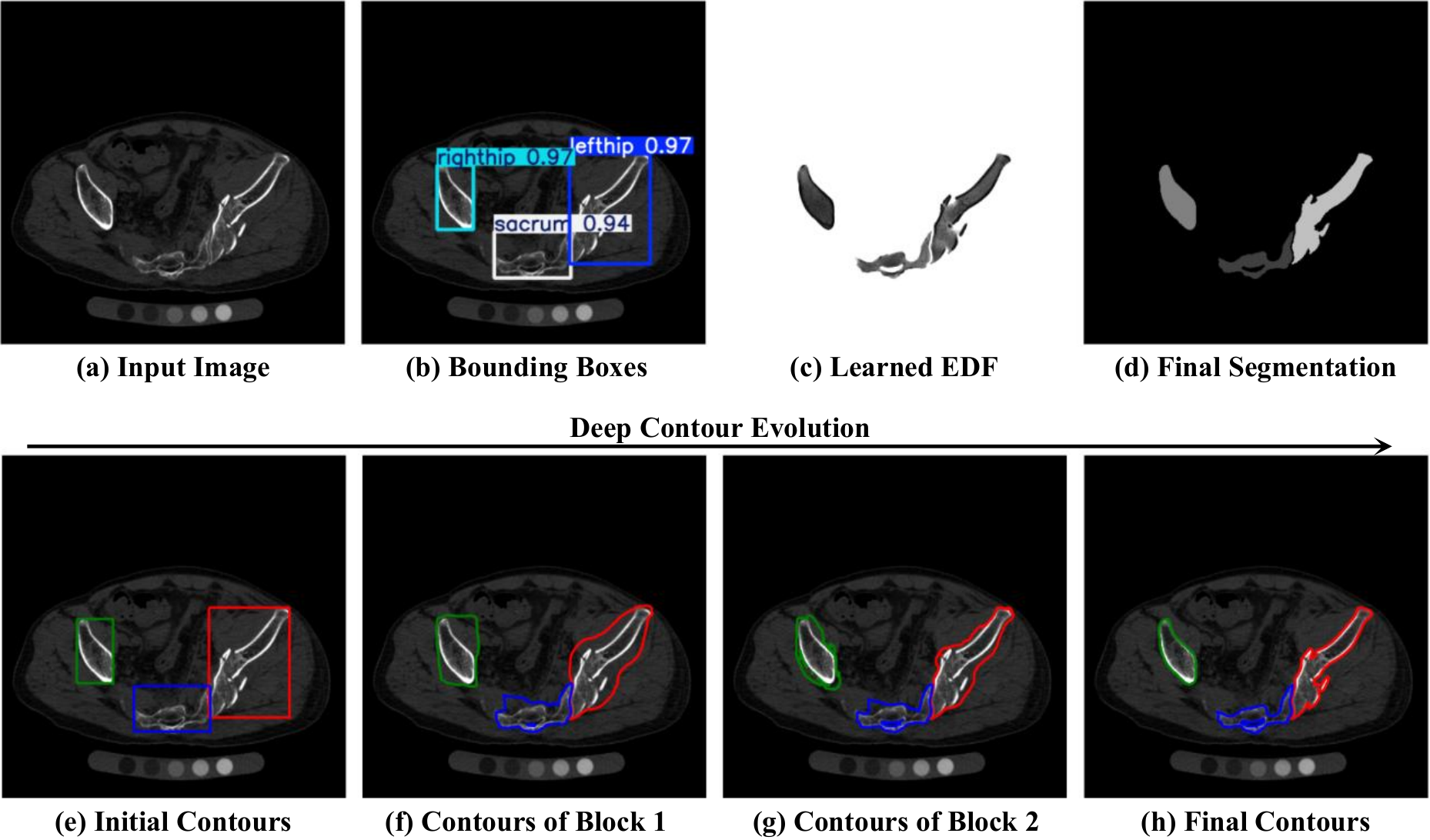}
\caption{The segmentation process of GUMP-Net from (a), (b), (c) to (e)-(h), and finally to (d). The decimal in (b) means the confidence of the predicted bounding box. From (e) to (h), the contours gradually evolve to the target boundaries.}
\label{wang1}
\end{figure}

The proposed algorithm not only provides an efficient segmentation method for complex fracture reduction, but also gives an interpretable geometric perspective for understanding deep learning segmentation, as illustrated in Fig. \ref{wang1}. 

\textit{\textbf{Outline.}} The paper is organized as follows. In section \ref{sec:preliminaries}, the required preliminaries are introduced. We propose the GUMP-Net method for multi-class pelvic segmentation in section \ref{sec:methods}. Section \ref{sec:experiments} gives the detailed experimental settings. In section \ref{sec:results}, numerical experiments and discussion are carried out to illustrate the performance of the proposed GUMP-Net. Finally, we conclude the paper in section \ref{sec:conclusion}. 

\section{Preliminaries}
\label{sec:preliminaries}
In this section, we briefly introduce some preliminaries related to this work, including classical GAC model, CNNs for image segmentation and object detection networks.

\subsection{Classical GAC Model}
\label{sec:gac}
Let $I : \Omega\subset\mathbb{R}^n \rightarrow \mathbb{R}$ $(n=2~\text{or}~3)$ be an image to be segmented. The level set function (LSF) $\phi$ is defined as a function on $\Omega$, which implicitly represents a contour by its zero-level set $\{x\in\Omega\,|\,\phi(x)=0\}$, and the topological changes of the contour can be automatically taken into account during the LSF evolution. In what follows, we adopt the convention that $\phi$ is negative inside the contour, which is the target region, and positive outside. 

The classical GAC model \cite{Caselles1995} aims to find a contour by minimizing its weighted length, and the corresponding level set formulation is
\begin{equation}
\label{eq1}
    \min_{\phi}\mathrm{Length}(\phi):=\int_{\Omega}g(|\nabla I(x)|)|\nabla H(\phi(x))|\mathrm{d}x,
\end{equation}
where $H$ denotes the Heaviside function and the weight is usually chosen as
\begin{equation*}
    g(|\nabla I(x)|)=\frac{1}{1+|\nabla I(x)|^2},
\end{equation*}
which takes small values when $x$ is located at image edges with large gradient norm. It encodes the image edge information and is often named edge detector function (EDF). It is noteworthy that this EDF will indiscriminately capture all image edges including the unconcerned ones, which might lead to undesired segmentation, especially in complex medical images with multiple anatomical structures. 

Practically, an area term is added into problem \eqref{eq1} to handle non-convex targets and speed up the contour evolution, which is defined by
\begin{equation*}
    \mathrm{Area}(\phi):=\alpha\int_{\Omega}g(|\nabla I(x)|)H(-\phi(x))\mathrm{d}x,
\end{equation*}
where $\alpha$ is a balancing weight. A positive (resp. negative) $\alpha$ results in contour shrinkage (expansion). The weight $\alpha$ should be carefully tuned in case the area term overwhelms the length term.

By introducing an artificial time into the LSF $\phi(t,\cdot\,)$, the associated gradient flow equation  is derived as
\begin{equation}\label{eq2}
\left\{
    \begin{aligned}
        &\frac{\partial\phi}{\partial t} = \left\{\nabla\cdot\left(g\frac{\nabla\phi}{|\nabla\phi|}\right)\!+\!\alpha g\right\}\delta(\phi),&&\text{in}\ (0,+\infty)\!\times\!\Omega,\\
        &\phi(0,x) = \phi_0(x),&&\text{in}\ \Omega,
    \end{aligned}\right.
\end{equation}
where $g$ denotes $g(|\nabla I|)$ for simplicity and $\delta(\,\cdot\,)$ stands for the Dirac delta function. The initial LSF $\phi_0$ can be chosen as the signed distance function with respect to the initial contour. Final segmentation is obtained by taking the negative region of the final LSF.

\subsection{CNNs for Image Segmentation}
The CNNs for image segmentation basically approximate the unknown mapping from the input image to the final segmentation by optimization on large amounts of labeled data. Among these, U-Net and its variants have been thoroughly investigated. U-Net \cite{Ronneberger2015} features a downsampling encoder, an upsampling decoder, and skip connections that enable direct information flow in-between and avoid possible information loss during downsampling. Variants like Attention U-Net \cite{Oktay2018} introduced attention gate (AG) before each scale of the decoder. This mechanism helps the network to focus on targets with varying shapes and sizes, such as the pancreas and the pelvis.

In the AG, each pixel of the feature map is assigned with an attention coefficient to highlight salient image regions and suppress irrelevant feature responses. Suppose that $\phi\in\mathbb{R}^{C_1\times W\times H}$ and $g\in\mathbb{R}^{C_2\times W\times H}$ are two input feature maps with $C_1$ and $C_2$ channels, and are then resized to $\phi=(\phi_i)\in\mathbb{R}^{C_1\times WH}$ and $g=(g_i)\in\mathbb{R}^{C_2\times WH}$, where $\phi_i\in\mathbb{R}^{C_1}$ and $g_i\in\mathbb{R}^{C_2}$ denote the column vector of $\phi$ and $g$ with respect to the $i$-th pixel, only in the context of the AG. The attention coefficient $\beta_i$ of the $i$-th pixel is computed by
\begin{equation*}
\beta_i=\mathrm{Sigmoid}(\psi^{\mathsf{T}}\mathrm{ReLU}(A_1\phi_i+A_2g_i+b_1)+b_2),
\end{equation*}
where $A_1\in\mathbb{R}^{F\times C_1}$, $A_2\in\mathbb{R}^{F\times C_2}$, $\psi\in\mathbb{R}^F$ are the linear transformations, $b_1\in\mathbb{R}^F$, $b_2\in\mathbb{R}$ are the bias terms and the superscript $F$ is a pre-determined hyper-parameter. Then the output of the AG is defined as
\begin{equation*}
    \omega=\mathrm{AG}_{\theta_{\mathrm{att}}}(\phi,g)=(\beta_i\phi_i)\in\mathbb{R}^{C_1\times WH},
\end{equation*}
where $\theta_{\mathrm{att}} = (A_1, A_2, b_1, b_2, \psi)$ denotes the network parameter, and then $\omega$ is resized again back to $\omega\in\mathbb{R}^{C_1\times W\times H}$. 

These CNNs were originally developed to predict the binary or multi-class segmentation masks, and have recently been further used to predict the signed distance functions (SDFs) \cite{Ma2021b}. The SDF contains not only the regional information, but also global distance information, which responses more sensitively to segmentation errors.

\subsection{Object Detection Networks}
Object detection focuses on localizing and classifying instances of objects from pre-defined semantic categories (e.g., humans, animals, or vehicles) in the images. The past two decades have witnessed a transfer in this area from hand-crafted methods to pure deep learning methods. Among these, a significant advancement is the emergence of highly efficient one-stage detectors. They frame the detection in a regression process where the location and classification are directly and simultaneously predicted. Representative examples are the YOLO model \cite{Redmon2016} and its subsequent versions, in which the full image is fed into a neural network to predict the bounding box information: the center coordinate $(a,b)$, the size $(w,h)$ and the class probability $p$. Beyond detection, the outputs of these models could benefit other downstream tasks. In our work, the bounding boxes are used as the initial contours in the active contour models, effectively providing spatial priors to capture fine structures and reduce segmentation outliers.

\section{Methods}
\label{sec:methods}
\begin{figure*}[htbp]
\includegraphics[height=6.62cm,width=12cm]{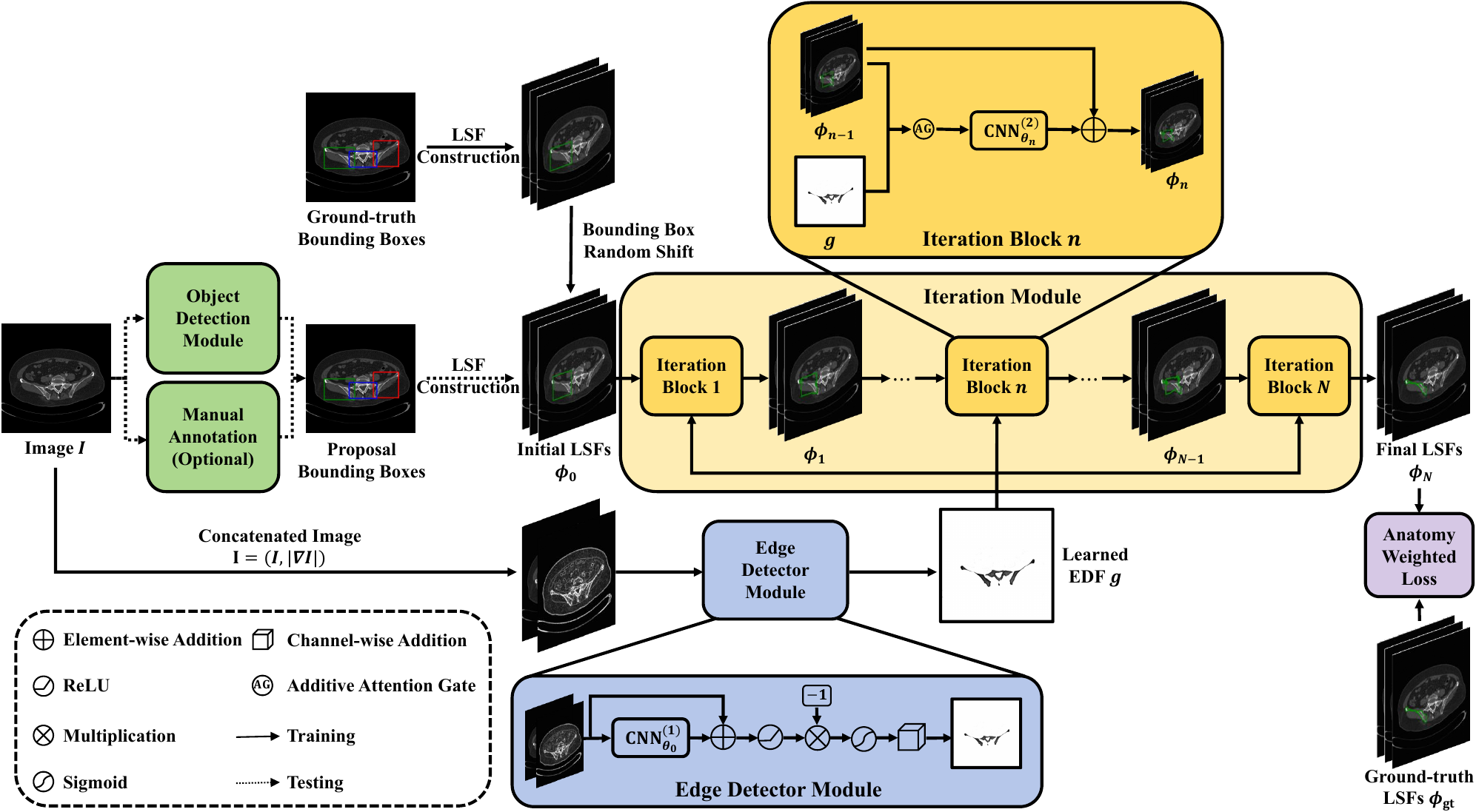}
\caption{The overall flowchart of the proposed GUMP-Net. The network architecture consists of three components: object detection module (ODM), edge detector module (EDM) and iteration module (IM). The ODM aims to facilitate automatic level set function (LSF) initialization. During network training, the initial LSFs are generated from the ground-truth bounding boxes, which are randomly shifted to reduce the dependency on the ODM and improve the robustness to initial contour deviation. The EDM aims to provide a clear learned edge detector function (EDF) for the deep contour evolution. The IM evolves the initial LSFs under the guidance of the learned EDF by $N$ consecutive CNN-based iteration blocks, so as to simulate the improved GAC iteration. Eventually, the final LSFs are supervised by the ground-truth LSFs with an anatomy weighted loss. In the implementation, $N$ is empirically taken as 3 for a trade-off between computational cost and accuracy.}
\label{wang2}
\end{figure*}

In this section, we detail the basic ingredients in the architecture of GUMP-Net\,: object dection module, edge detector module and iteration module. Then we summarize the overall flowchart in Fig. \ref{wang2}, and introduce how GUMP-Net trains and runs.

\subsection{Object Detection Module}
The object detection module (ODM) aims to automatize the initialization of the contour or the level set function (LSF) evolution and reduce laborious interactions. A \textit{pre-trained} one-stage object detector can be utilized to provide the bounding boxes as initial contours of the concerned anatomies, such as the \textit{left hip}, \textit{right hip}, \textit{sacrum} in the pelvic CT images and \textit{fibula}, \textit{tibia} in the ankle CT images. Alternatively, if the results of the ODM are not trusted by the clinicians, the bounding boxes can still be given by manual annotations.

After the object detection, each anatomy should be assigned with an initial LSF. All initial LSFs of the concerned anatomies combine together to form a multi-channel initial LSF, and this procedure is named ``LSF Construction'' in Fig. \ref{wang2}. Mathematically, for a certain anatomy $c$, such as the right hip, the rectangular region $B^c$ of the bounding box is determined by its center $(a,b)$ and size $(w,h)$, and its boundary is denoted as $\partial B^c$. Then, the corresponding initial LSF can be taken as the SDF, i.e.,
\begin{equation*}
    \phi_0^c(x)=\mathrm{SDF}(x;B^c)=\left\{
    \begin{aligned}
        &+d(x,\partial B^c),&& x\notin B^c,\\
        &-d(x,\partial B^c),&& x\in B^c,
    \end{aligned}\right.
\end{equation*}
where $d(x,\partial B^c)$ is the Euclidean distance from $x$ to $\partial B^c$. If the anatomy $c$ does not appear in the image, the initial LSF is chosen as $\phi_0^c(x)=d(x,O)$ where $O$ is the center of the image. Finally, the multi-channel initial LSF is therefore constructed as
\begin{equation*}
    \bm{\phi}_0=(\phi_0^1,\,\cdots,\phi_0^C),
\end{equation*}
which is realized by channel-wise concatenation.

\subsection{Edge Detector Module}
The edge detector module (EDM) is based on the mathematical form of an improved edge detector function (EDF) and aims to learn a clear EDF for the deep contour evolution. As mentioned in the subsection \ref{sec:gac}, the EDF in the classical GAC model indiscriminately captures all image edges with large gradient norm, which might influence the contour evolution and lead to segmentation errors. To overcome this, fracture interactive GAC (FI-GAC) \cite{WANG2026113049} took medical knowledge into account and designed a novel EDF that can suppress soft-tissue responses and emphasize bone features, whose formula is as follows:
\begin{equation}\label{eq3}
    g(I,|\nabla I|)\!=\!\frac{1}{1+f_1(\sigma(I-r_1))}+\frac{\gamma}{1+f_2(\sigma(|\nabla I|-r_2))},
\end{equation}
where $\sigma(z)=\max(z,0)$ is the ReLU activation function, $f_1$ and $f_2$ are two high-pass filters to enhance the contrast, $\gamma$ is a weight to balance two components. It can be observed in \eqref{eq3} that the gray-scale values smaller than $r_1$ will be truncated to zero by the ReLU and those larger than $r_1$ will be preserved as $I-r_1$. This means the soft-tissue with smaller gray-scale value will be suppressed and the bone with relatively larger gray-scale value will be highlighted. This observation also holds similarly for the gradient norm and thus weak edges will be eliminated. Therefore, the key to distinguishing bone from soft-tissue lies in the thresholds $r_1$ and $r_2$, which can be derived via two optimization problems constrained by the orthopedic knowledge, such as the range of the CT value and that of the CT window. 

However, this hand-crafted design relies on the pre-defined range of the CT window and the selected thresholds. On the other hand, the selected thresholds $r_1$ and $r_2$ are constant for all pixels, which might be rigid and does not consider the location information. A natural idea comes to learn this orthopedic knowledge automatically from data. To endow the thresholds with more flexibility, they are considered to vary with the pixel, that is to be, $r_1(x)$ and $r_2(x)$, which can be predicted by a CNN termed $\mathrm{CNN_{\theta_0}^{(1)}}$ according to the image $I$ and the gradient norm $|\nabla I|$, formally,
\begin{equation}\label{eq4}
    \mathbf{r}=(-r_1,-r_2)=\mathrm{CNN}_{\theta_0}^{(1)}(\mathbf{I}),
\end{equation}
where $\mathbf{I}=(I,|\nabla I|)$ is the concatenated image and $\theta_0$ is the network parameter. Next, take $f_1(z)=f_2(z)=\mathrm{exp}(z)$ in \eqref{eq3} and construct an EDF accordingly,
\begin{equation}\label{eq5}
    \mathbf{p}=(p_1,p_2)=\mathrm{Sigmoid}(-\mathrm{ReLU}(\mathbf{I}+\mathbf{r})),
\end{equation}
\begin{equation}\label{eq6}
     g = p_1 + \gamma p_2,
\end{equation}
where $\gamma$ can be either fixed or learned. Without loss of generality, we set $\gamma=1$. Hence, \eqref{eq4}-\eqref{eq6} constitute the proposed EDM to automatically provide a learned EDF $g$ from the concatenated image $\mathbf{I}$, as shown in Fig. \ref{wang2}.

\subsection{Iteration Module}
The iteration module (IM) aims to evolve the multi-channel initial level set function (LSF) to obtain the final LSF under the guidance of the learned edge detector function (EDF). It consists of $N$ consecutive iteration blocks that employ neural networks to simulate the improved GAC iterations. A simple iteration scheme of \eqref{eq2} can be derived by forward Euler method:
\begin{equation}\label{eq7}
\phi_n=\phi_{n-1} + h\left\{\nabla\cdot\left(g\frac{\nabla\phi_{n-1}}{|\nabla\phi_{n-1}|}\right)+\alpha g\right\} \delta(\phi_{n-1}),
\end{equation}
where $h$ is the time step size. The second item on the right-hand side (or the residual term) depends on the EDF $g$ and the previous LSF $\phi_{n-1}$, which inspires the idea to predict the residual term from $g$ and $\phi_{n-1}$ by a neural network $\mathrm{NN}_{\theta_{n}}$. Therefore, \eqref{eq7} becomes
\begin{equation}\label{IM1}
    \bm{\phi}_n = \bm{\phi}_{n-1} + \mathrm{NN}_{\theta_n}(\bm{\phi}_{n-1},g),\quad n=1,\ldots,N, 
\end{equation}
where $\bm{\phi}_n$ is the multi-channel LSF and $\theta_n$ is the network parameter of the $n$-th step. For the design of the neural network $\mathrm{NN}_{\theta_{n}}$, in our implementation, an additive attention gate $\mathrm{AG}_{\theta_{n}}$ is utilized at first to help the previous LSF $\bm{\phi}_{n-1}$ focus on the learned EDF $g$, thereby incorporating both of the information. Next, another CNN termed $\mathrm{CNN}_{\theta_{n}}^{(2)}$ is employed to predict the residual term from the incorporated information, i.e.,
\begin{equation}\label{IM2}
    \mathrm{NN}_{\theta_n}(\bm{\phi}_{n-1},g) = \mathrm{CNN}_{\theta_{n}}^{(2)}(\mathrm{AG}_{\theta_{n}}(\bm{\phi}_{n-1},g)).
\end{equation}
Then, \eqref{IM1} and \eqref{IM2} combine together to form the proposed IM, as illustrated in Fig. \ref{wang2}.

\subsection{Network Training}
In this subsection, we detail the level set function (LSF) initialization during network training and the loss functions, and present the training algorithm of GUMP-Net in Algorithm \ref{Al1}.

\subsubsection{LSF Initialization during Training}
During the network training, an initial multi-channel LSF is required for the deep contour evolution. Unlike the inference process where initial contours are predicted by the object detection module (ODM), the ground-truth masks are available during training and can be naturally used to generate tight ground-truth bounding boxes by their extreme points. 

Furthermore, in order to reduce the dependency on the performance of the ODM during inference and improve the robustness to possible initial contour misalignment, we propose a training technique called ``Bounding Box Random Shift''. To be detailed, the initial LSF is constructed according to the ground-truth bounding box randomly shifted to a small degree $\varepsilon$. Formally, for a certain anatomy $c$, the shifted initial LSF is computed by
\begin{equation*}
    \phi_{0,\varepsilon}^c(x)=\mathrm{SDF}(x;\mathcal{R}_\varepsilon(B^{c})),
\end{equation*}
where $B^c$ denotes the ground-truth bounding box here, $\mathcal{R}_\varepsilon$ is the random shift operation and the shifted bounding box $\mathcal{R}_\varepsilon(B^{c})$ is determined by the shifted center $(a_\varepsilon,b_\varepsilon)$ and the shifted size $(w_\varepsilon,h_\varepsilon)$, as follows:
\begin{equation*}
\begin{aligned} 
    (a_\varepsilon,b_\varepsilon) &= (a + \epsilon_1, b + \epsilon_2),\\
    (w_\varepsilon,h_\varepsilon) &= (w + \epsilon_3, h + \epsilon_4),
\end{aligned}
\end{equation*}
where $\{\epsilon_j\}_{j=1}^4$ are independently sampled from the uniform distribution on $[-\varepsilon,\varepsilon]$. In our implementation, $\varepsilon$ is set as 10 pixels. 
The shifted multi-channel initial LSF is then defined as
\begin{equation*}
    \bm{\phi}_{0,\varepsilon}=(\phi_{0,\varepsilon}^1,\,\cdots,\phi_{0,\varepsilon}^C),
\end{equation*}
which is used in the training procedure.

\subsubsection{Training Loss}
During the network training, three types of losses are jointly optimized: the binary cross entropy (BCE) loss $L_{\mathrm{BCE}}$, the Dice loss $L_{\mathrm{Dice}}$ and the averaged LSF loss $L_{\mathrm{LSF}}$. 

For the final LSF $\phi_N^c$ of the anatomy $c$, the corresponding total loss $L$ is computed by
\begin{equation*}
\begin{aligned}
L(\phi_N^c) &= w_1L_{\mathrm{BCE}}(S_{\mathrm{gt}}^c,S_{\mathrm{pred}}^c) + w_2L_{\mathrm{Dice}}(S_{\mathrm{gt}}^c,S_{\mathrm{pred}}^c) + w_3L_{\mathrm{LSF}}(\phi_{\mathrm{gt}}^c,\phi_N^c), 
\end{aligned}
\end{equation*}
where $S_{\mathrm{gt}}^c$ is the characteristic function of the ground-truth mask, $S_{\mathrm{pred}}^c=\mathrm{Sigmoid}(-k\phi_N^c)$ is the corresponding predicted probabilistic function, $\phi_{\mathrm{gt}}^c$ is the corresponding ground-truth LSF defined as the SDF with respect to the ground-truth mask. The averaged LSF loss is computed by
\begin{equation*}
    L_{\mathrm{LSF}}(\phi_{\mathrm{gt}}^c,\phi_N^c)=\frac{1}{|\Omega|}\sum_{x\in\Omega}|\phi_{\mathrm{gt}}^c(x) - \phi_N^c(x)|,
\end{equation*}
where $|\Omega|$ is the number of pixels in $\Omega$. Without loss of generality, here the balancing weights are taken as $w_1=10,w_2=1,w_3=0.01$ and the parameter $k$ is set to be 1500 in the following experiments. 

Next, an anatomy weighted loss ($AWL$) is designed by summing the losses of all anatomies according to their segmentation complexity:
\begin{equation*}
    AWL(\bm{\phi}_N;\Theta)=\sum_{c=1}^C\lambda_c L(\phi_N^c),
\end{equation*}
where $\Theta=(\theta_0,\,\cdots,\theta_N)$ is the network parameter of the whole architecture, $\lambda_c$ is the weight for the anatomy $c$. For example, in the segmentation of the pelvic CT images, the left hip and right hip are relatively easier to segment, while the sacrum is much more difficult due to its small size and various shape which should be focused with more attention. Therefore, in our implementation, we weight the left hip, right hip and sacrum by $0.25$, $0.25$ and $0.5$ respectively.

\begin{algorithm}
\caption{GUMP-Net Training Algorithm}
\label{Al1}
\begin{algorithmic}[1]
\Require Image $I$, multi-channel ground-truth mask $\mathbf{S}_{\mathrm{gt}}=(S_{\mathrm{gt}}^1,\,\cdots,S_{\mathrm{gt}}^C)$ and its corresponding randomly shifted initial multi-channel LSF $\bm{\phi}_{0,\varepsilon}$. Initial network parameter $\Theta$, learning rate $\xi$ and parameters $\gamma$ and $\varepsilon$. 
\State $\mathbf{I}\leftarrow(I,|\nabla I|)$\hfill$\blacktriangleright$ Edge Detector Module
\State $\mathbf{r}\leftarrow\mathrm{CNN}_{\theta_0}^{(1)}(\mathbf{I})$
\State $(p_1,p_2)\leftarrow\mathrm{Sigmoid}(-\mathrm{ReLU}(\mathbf{I}+\mathbf{r}))$
\State $g\leftarrow p_1+\gamma p_2$
\State $\bm{\phi}_0\leftarrow\bm{\phi}_{0,\varepsilon}$\hfill$\blacktriangleright$ Level Set Initialization
\For{$n\leftarrow 1,\ldots,N$}\hfill$\blacktriangleright$ Iteration Module
    \State $\bm{\phi}_n \leftarrow \bm{\phi}_{n-1} + \mathrm{CNN}_{\theta_{n}}^{(2)}(\mathrm{AG}_{\theta_{n}}(\bm{\phi}_{n-1},g))$
\EndFor
\State $\Theta\leftarrow\Theta-\xi\frac{\partial}{\partial\Theta}AWL(\bm{\phi}_N;\Theta)$
\State Resample $I$, $\mathbf{S}_{\mathrm{gt}}$ and $\bm{\phi}_{0,\varepsilon}$ and go to step $1$. Repeat this loop until the network converges.
\end{algorithmic}
\end{algorithm}

\section{Experimental Settings}
\label{sec:experiments}
In this section, we elaborate the datasets, the implementation details and the evaluation metrics. 

\subsection{Datasets}
We implement our method on several datasets: open-sourced CTPelvic1K dataset6 CLINIC (Pelvic1K CLINIC) \cite{Liu2020}, pelvic CT dataset (Pelvic Collected) and ankle CT dataset (Ankle Collected) collected by collaborative hospitals. Following is a brief description of the datasets.

\begin{itemize}
    \item \textbf{Pelvic1K CLINIC.} This dataset includes preoperative images with pelvic fractures. The average size of all volumes is $(512,512,345)$, with a resolution of $0.85\times 0.85\times 0.8\ \mathrm{mm}^3$.
    \item \textbf{Pelvic Collected.} This dataset includes pelvic CT images without fractures, in which half of the volumes especially annotate the coccyx. To ensure data consistency, we only use these volumes with annotated coccyx. The average size of all volumes is $(512,512,190)$, with a resolution of $0.8\times 0.8\times 1.2\ \mathrm{mm}^3$.
    \item \textbf{Ankle Collected.} This dataset includes ankle CT images with annotated distal tibia and fibula. The average size of all volumes is $(768,768,260)$, with a resolution of $0.3\times 0.3\times 0.4\ \mathrm{mm}^3$.
\end{itemize}

\subsection{Implementation Details}
\subsubsection{Architecture}
For the object detection module (ODM), we utilize the YOLO framework\footnote{Available at: https://docs.ultralytics.com/}, which is a relatively simple yet efficient one-stage object detector. Considering that medical images usually contain fewer and more limited semantic categories than the real-world images, we choose the YOLOv8s model for a fast detection speed. The detector is individually pre-trained on each dataset for 200 epochs. The dataset splits and the detection metrics are reported in Table \ref{Tab1}. Optionally, more recent versions such as YOLO11 and YOLO26 can also be used in the architecture.

\begin{table}[h!]
\centering
\caption{Dataset split and detection metric during detector pre-training. `\#' represents the number of data volumes. `$\mathrm{Tr/Ts}$' denotes training/testing set. The metric mAP@0.5 is the mean average precision across multiple categories at the confidence of 0.5.}
\vspace{3mm}
\label{Tab1}
\begin{tabular}{cccc}
\toprule
Datasets & Pelvic1K CLINIC & Pelvic Collected & Ankle Collected\\  
\midrule
\# of Tr/Ts      & 30/6                     & 50/10                     & 28/4                    \\
mAP@0.5          & 98.1\%                   & 98.8\%                    & 99.5\%                  \\
\bottomrule
\end{tabular}
\end{table}

For the edge detector module (EDM), to better capture fine features and bone structures, we employ an Attention U-Net which downsamples four times with intermediate channels \{32,64,128,256,512\}  as the $\mathrm{CNN}^{(1)}_{\theta_0}$, around 8M parameters. 

For the iteration module (IM), to facilitate multiple iteration blocks, we employ a relatively smaller U-Net which also downsamples four times with intermediate channels \{16,32,64,128,256\} as the $\mathrm{CNN}^{(2)}_{\theta_n}$, around 2M parameters. In the implementation, we empirically take $N=3$ iteration blocks for a trade-off between computational cost and accuracy.

\subsubsection{Data Augmentation} On Pelvic1K CLINIC and Pelvic Collected, we only apply small angle $(\pm 10^\circ)$ random rotation to simulate the patient posture variations. Horizontal flipping is not applied in case of possible network confusion, since the left hip and the right hip are almost symmetric. On Ankle Collected, we apply both small angle $(\pm 15^\circ)$ random rotation and horizontal flipping to simulate various leg postures and balance the data from the left and right legs. All images are resized to $512\times 512$ and normalized to $[0,1]$ before network training.

\subsubsection{Training Parameters} All the experiments are implemented with Python 3.9 and PyTorch 2.5 and are run in Ubuntu 22.04 with a NVIDIA A5000 24GB GPU. The proposed network is trained by Adam optimizer with default parameters for 150 epochs. The batch size is taken as $8$. The learning rate is set with an initial value of $10^{-4}$ and then decays for the last 25 epochs following the schedule in \cite{Isensee2021}. 

\subsection{Evaluation Metrics}
We use four metrics to evaluate the segmentation performance. In what follows, $G$ and $S$ represent the ground-truth and the segmentation result respectively, $\partial(\,\cdot\,)$ means the boundary and $|\cdot|$ means the Hausdorff measure.
\begin{itemize}
    \item Dice and Jaccard coefficient:
    \begin{equation*}
        \mathrm{Dice}=2\frac{|G\cap S|}{|G|+|S|},\ 
        \mathrm{Jaccard}=\frac{|G\cap S|}{|G\cup S|}.
    \end{equation*}
    They evaluate the overlapping ratio between the ground-truth and the segmentation mask. $100\%$ indicates a perfect segmentation.
    \item Hausdorff distance (HD) and average symmetric surface distance (ASSD):
    \begin{equation*}
        \mathrm{HD} = \max\left\{\max_{x\in\partial G}d(x,\partial S),\max_{y\in\partial S}d(\partial G,y)\right\},
    \end{equation*}
    \begin{equation*}
        \mathrm{ASSD} = \frac{\sum_{x\in\partial G}d(x,\partial S)+\sum_{y\in\partial S}d(\partial G,y)}{|\partial G|+|\partial S|}.
    \end{equation*}
    They evaluate the deviation between the boundary of the ground-truth and that of the segmentation mask. Zero value indicates a perfect segmentation.
\end{itemize}

\section{Experimental Results and Discussion}
\label{sec:results}
In this section, we present and discuss the results of the following three groups of experiments in detail to demonstrate the effectiveness of the proposed GUMP-Net. 

\subsection{Ablation Study of the EDM and ODM} 
\label{Ablation}
We first examine the necessity and effectiveness of the proposed edge detector module (EDM) and object detection module (ODM) through ablation studies across all three datasets. The dataset splits for Pelvic1K CLINIC, Pelvic Collected and Ankle Collected are 16/2, 24/3 and 16/2 respectively. 

\begin{figure}[htbp]
\centering
\includegraphics[height=5.64cm,width=10cm]{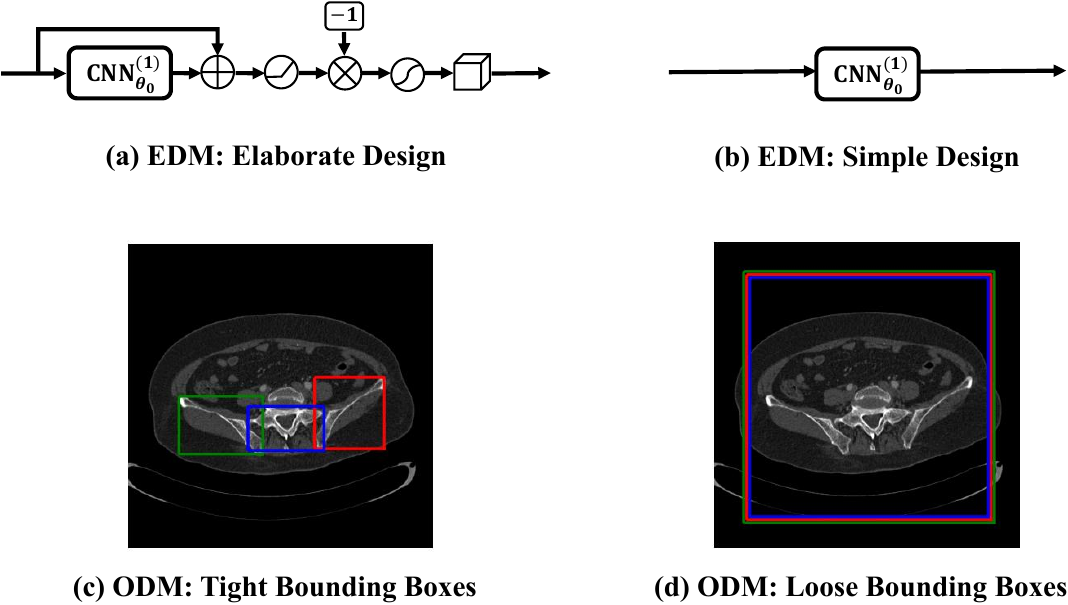}
\caption{An illustration of the ablation study of the proposed edge detector module (EDM) and object detection module (ODM).}
\label{wang5}
\end{figure}

For the ablation study of the EDM, we compare the elaborate design in Fig. \ref{wang5}(a) with a simple design in Fig. \ref{wang5}(b). Both designs share the same CNN backbone, but the elaborate one incorporates structural information from the formulation of FI-GAC. Quantitative results in Table \ref{Tab2} show that for all three datasets the elaborate EDM design achieves most of the best scores, particularly in sacrum and tibia. 

For the ablation study of the ODM, we train our network under two kinds of initialization strategies: (1) tight bounding boxes as Fig. \ref{wang5}(c); (2) loose and fixed size bounding boxes as Fig. \ref{wang5}(d). Corresponding results in Table \ref{Tab2} show that tight bounding boxes consistently improve the performance, especially for small structures like sacrum. The standard deviation is also reduced, indicating a more robust segmentation. Therefore, it is necessary to provide tight bounding boxes when available, for the sake of better segmentation performance. 

\begin{table}[h!]
\centering
\renewcommand{\arraystretch}{1.3}
\caption{Ablation study of the proposed edge detector module (EDM) and object detection module (ODM) on three datasets. For the EDM, we have (a) elaborate design and (b) simple design, and for the ODM, we propose (c) tight bounding boxes and (d) loose bounding boxes, as shown in Fig. \ref{wang5}. Left hip and right hip are abbreviated to left and right respectively for brevity. Bold font means the best among the compared cases. }
\vspace{3mm}
\label{Tab2}
\begin{adjustbox}{max width=\columnwidth, center}
\begin{tabular}{cc|c|c|c|c|c}
\hline
Datasets                           & \makecell{EDM+ODM}           & Anatomies   & Dice (\%) $\uparrow$  & Jaccard (\%) $\uparrow$ & HD $\downarrow$ & ASSD $\downarrow$ \\ \hline
\multirow{9}{*}{\makecell{Pelvic1K\\CLINIC}}  & \multirow{3}{*}{\makecell{(a)+(c)}} & Left & \textbf{98.68$\pm$3.29}  & 97.56$\pm$4.90    & \textbf{2.13$\pm$3.81}  & \textbf{0.12$\pm$0.18}    \\
                                  &                        & Right & 98.97$\pm$4.18  & 98.19$\pm$4.89    & 2.69$\pm$7.47  & 0.13$\pm$0.34    \\
                                  &                        & Sacrum    & \textbf{97.51$\pm$6.10}  & \textbf{95.66$\pm$8.80}    & \textbf{1.63$\pm$3.26}  & \textbf{0.14$\pm$0.22}    \\ \cline{2-7} 
                                  & \multirow{3}{*}{\makecell{(b)+(c)}}    & Left  & 98.57$\pm$7.08 & 97.70$\pm$7.54    & 3.67$\pm$7.85  & 0.23$\pm$0.52    \\
                                  &                           & Right & \textbf{99.00$\pm$4.23} & \textbf{98.23$\pm$4.93}    & \textbf{1.44$\pm$3.80}  & \textbf{0.09$\pm$0.12}    \\
                                  &                           & Sacrum    & 97.31$\pm$8.60 & 95.60$\pm$10.36    & 1.70$\pm$3.43  & 0.15$\pm$0.23    \\ \cline{2-7}
                                  & \multirow{3}{*}{\makecell{(a)+(d)}} & Left  & 98.68$\pm$5.49  & \textbf{97.71$\pm$5.92}    & 2.78$\pm$8.34  & 0.13$\pm$0.19    \\
                                  &                        & Right & 98.62$\pm$5.87  & 97.64$\pm$6.60    & 1.83$\pm$4.77  & 0.13$\pm$0.22    \\
                                  &                        & Sacrum    & 96.71$\pm$9.84  & 94.70$\pm$11.57   & 4.09$\pm$11.04 & 0.28$\pm$0.89    \\ \hline
\multirow{9}{*}{\makecell{Pelvic\\Collected}} & \multirow{3}{*}{\makecell{(a)+(c)}} & Left  & \textbf{96.96$\pm$4.11}  & 94.31$\pm$5.56    & 2.38$\pm$3.39  & 0.39$\pm$0.43    \\
                                  &                        & Right & \textbf{96.64$\pm$3.79}  & \textbf{93.73$\pm$6.20}    & 2.86$\pm$3.55  & 0.42$\pm$0.36    \\
                                  &                        & Sacrum    & \textbf{95.57$\pm$7.05}  & \textbf{92.18$\pm$10.25}   & \textbf{2.63$\pm$4.00}  & \textbf{0.37$\pm$0.44}    \\ \cline{2-7}
                                  & \multirow{3}{*}{\makecell{(b)+(c)}}    & Left  & 96.69$\pm$4.16 & \textbf{94.32$\pm$5.69}    & \textbf{2.33$\pm$2.46}  & 0.39$\pm$0.38    \\
                                  &                           & Right & 96.30$\pm$5.33 & 93.27$\pm$7.99    & 2.99$\pm$4.62  & 0.43$\pm$0.44    \\
                                  &                           & Sacrum    & 95.18$\pm$9.44 & 91.83$\pm$11.84   & 2.83$\pm$4.69  & 0.40$\pm$0.60    \\ \cline{2-7}
                                  & \multirow{3}{*}{\makecell{(a)+(d)}} & Left  & 96.46$\pm$8.72  & 93.95$\pm$9.50    & 2.50$\pm$2.68  & \textbf{0.38$\pm$0.49}    \\
                                  &                        & Right & 96.09$\pm$9.77  & 93.42$\pm$10.19   & \textbf{2.74$\pm$3.00}  & \textbf{0.38$\pm$0.27}    \\
                                  &                        & Sacrum    & 92.58$\pm$15.59 & 88.70$\pm$17.81   & 4.11$\pm$9.23  & 0.71$\pm$2.37    \\ \hline
\multirow{6}{*}{\makecell{Ankle\\Collected}}  & \multirow{2}{*}{\makecell{(a)+(c)}} & Tibia     & \textbf{99.48$\pm$0.91}  & \textbf{98.99$\pm$1.70}    & \textbf{0.96$\pm$1.05}  & \textbf{0.14$\pm$0.12}    \\
                                  &                        & Fibula    & \textbf{98.62$\pm$1.89}  & \textbf{97.33$\pm$3.13}    & 2.49$\pm$3.74  & \textbf{0.25$\pm$0.29}    \\ \cline{2-7}
                                  & \multirow{2}{*}{\makecell{(b)+(c)}}    & Tibia     & 99.36$\pm$1.78 & 98.78$\pm$2.78    & 1.07$\pm$1.36  & 0.15$\pm$0.11    \\
                                  &                           & Fibula    & 98.34$\pm$2.32 & 96.82$\pm$3.62    & 2.47$\pm$3.70  & 0.29$\pm$0.30    \\ \cline{2-7}
                                  & \multirow{2}{*}{\makecell{(a)+(d)}} & Tibia     & 99.13$\pm$4.80  & 98.52$\pm$5.22    & 1.08$\pm$1.92  & 0.15$\pm$0.11    \\
                                  &                        & Fibula    & 98.19$\pm$2.55  & 96.55$\pm$3.75    & \textbf{2.34$\pm$3.45}  & 0.29$\pm$0.26    \\ \hline
\end{tabular}
\end{adjustbox}
\end{table}

\subsection{Comparison with State-of-the-art Methods} 
We compare GUMP-Net with several state-of-the-art methods from both quantitative and visual aspects, including CNN-based benchmarks (DeepLabv3+ \cite{Chen2018}, U-Net \cite{Ronneberger2015}, Attention U-Net \cite{Oktay2018}, nnU-Net \cite{Isensee2021}), Transformer-based Swin-Unet \cite{Hu2021}, algorithm unrolling methods (FAS-UNet \cite{Zhang2022}, PottsMGNet \cite{Tai2024}) and foundation models in image segmentation (SAM \cite{SAM}, MedSAM \cite{MEDSAM}). The compared methods are either well-established or up-to-date. 

In the quantitative comparison, it is unpractical to compare with SAM and MedSAM due to massive slice-by-slice interactions. As for other methods, to be fair, we retrain these methods with default settings and parameters. The dataset splits for Pelvic1K CLINIC, Pelvic Collected and Ankle Collected are 24/6, 32/8 and 16/4 respectively. As shown in Table \ref{Tab5}, the proposed GUMP-Net achieves the best or the second best scores across all metrics. On average, GUMP-Net exceeds the second best in Dice and Jaccard by 0.5-1\%, in HD by 0.5 and in ASSD by 0.05, which is particularly obvious for sacrum. 
\begin{table}[htbp]
\centering\tiny
\renewcommand{\arraystretch}{1}
\caption{Evaluation metrics of different methods on three datasets. Left hip and right hip are abbreviated to left and right respectively for brevity. Bold font means the best score among all compared methods, and the underlined score means the second best. }
\vspace{3mm}
\label{Tab5}
\begin{tabular}{cl|p{0.3cm}p{0.3cm}p{0.4cm}|p{0.3cm}p{0.3cm}p{0.4cm}|p{0.3cm}p{0.3cm}p{0.4cm}|p{0.3cm}p{0.3cm}p{0.4cm}}
\hline
Datasets                           & \multicolumn{1}{l|}{Methods}          & \multicolumn{3}{c|}{Dice (\%) $\uparrow$}      & \multicolumn{3}{c|}{Jaccard (\%) $\uparrow$}   & \multicolumn{3}{c|}{HD $\downarrow$}        & \multicolumn{3}{c}{ASSD $\downarrow$}      \\ \hline
                                  &    Anatomies             & Left & Right & Sacrum & Left & Right & Sacrum & Left & Right & Sacrum & Left & Right & Sacrum \\ \hline
\multirow{8}{*}{\makecell{Pelvic1K\\CLINIC}}  & DeepLabv3+      & 93.37    & 92.85     & 87.78  & 90.59    & 90.42     & 83.84  & 10.60    & 5.69      & 6.26   & 0.75     & 0.48      & 1.00   \\
                                  & U-Net           & \underline{98.67}    & 98.11     & 93.20  & \underline{97.77}    & 97.17     & 91.52  & 2.76     & 3.31      & 2.95   & 0.20     & 0.25      & 0.41   \\
                                  & Attention U-Net & 97.04    & \underline{98.41}     & \underline{95.64}  & 95.94    & \underline{97.32}     & \underline{93.89}  & 3.58     & 2.08      & \underline{2.90}   & 0.43     & \underline{0.15}      & \underline{0.32}   \\
                                  & nnU-Net         & 97.01    & 96.18     & 92.20  & 95.89    & 95.10     & 90.33  & 3.71     & 3.90      & 5.38   & 0.46     & 0.26      & 0.39   \\
                                  & Swin-Unet       & 94.45    & 92.54     & 88.66  & 90.90    & 88.84     & 84.10  & 3.92     & 5.00      & 5.85   & 0.46     & 0.71      & 0.98   \\
                                  & FAS-UNet        & 95.81    & 97.05     & 92.79  & 94.81    & 96.15     & 90.85  & 6.53     & 3.33      & 4.34   & 1.20     & 0.17      & 0.58   \\
                                  & PottsMGNet      & 98.06    & 95.71     & 93.07  & 97.15    & 94.84     & 91.18  & \underline{2.40}     & \textbf{2.05}      & 4.10   & \textbf{0.12}     & 0.18      & 0.57   \\
                                  & \textbf{GUMP\,(Ours)}      & \textbf{99.03}    & \textbf{99.01}     & \textbf{97.64}  & \textbf{98.17}    & \textbf{98.20}     & \textbf{96.05}  & \textbf{1.96}     & \underline{2.07}      & \textbf{1.75}   & \underline{0.13}     & \textbf{0.12}      & \textbf{0.18}   \\ \hline
\multirow{8}{*}{\makecell{Pelvic\\Collected}} & DeepLabv3+      & 95.91    & 96.41     & 88.98  & 93.00    & 93.56     & 84.21  & 5.08     & 4.61      & 5.77   & 0.52     & 0.48      & 1.06   \\
                                  & U-Net           & 97.02    & 96.60     & \underline{94.23}  & 94.78    & 94.07     & \underline{90.80}  & 2.84     & 2.81      & \underline{4.09}   & 0.35     & 0.39      & \underline{0.59}   \\
                                  & Attention U-Net & \underline{97.49}    & \underline{96.94}     & 93.40  & \underline{95.64}    & \underline{94.73}     & 89.84  & 2.33     & 2.38      & 4.13   & \textbf{0.27}     & \textbf{0.32}      & 0.70   \\
                                  & nnU-Net         & 97.14    & 96.78     & 93.60  & 94.99    & 94.25     & 90.17  & \textbf{2.11}     & \textbf{2.33}      & 4.23   & 0.31     & 0.36      & 0.84   \\
                                  & Swin-Unet       & 94.20    & 94.12     & 88.82  & 89.83    & 89.89     & 83.65  & 4.28     & 4.76      & 5.52   & 0.63     & 0.64      & 1.09   \\
                                  & FAS-UNet        & 97.13    & 96.84     & 91.65  & 95.10    & 94.58     & 87.51  & 3.04     & 3.02      & 5.13   & 0.32     & 0.34      & 0.90   \\
                                  & PottsMGNet      & 96.99    & 96.75     & 92.63  & 94.89    & 94.51     & 88.96  & 2.64     & 3.22      & 4.20   & 0.32     & 0.45      & 0.71   \\
                                  & \textbf{GUMP\,(Ours)}      & \textbf{97.71}    & \textbf{97.38}     & \textbf{95.25}  & \textbf{95.70}    & \textbf{95.05}     & \textbf{91.98}  & \underline{2.19}     & \underline{2.36}      & \textbf{2.79}   & \underline{0.29}     & \underline{0.33}      & \textbf{0.42}   \\ \hline
                                  &   Anatomies              & Tibia    & Fibula    &        & Tibia    & Fibula    &        & Tibia    & Fibula    &        & Tibia    & Fibula    &        \\ \hline
\multirow{8}{*}{\makecell{Ankle\\Collected}}  & DeepLabv3+      & 87.40    & 92.04     &        & 85.10    & 88.49     &        & 9.06     & 6.00      &        & 1.17     & 1.15      &        \\
                                  & U-Net           & 97.47    & 96.35     &        & 96.70    & 94.78     &        & 1.83     & 3.25      &        & 0.23     & 0.56      &        \\
                                  & Attention U-Net & \underline{98.16}    & 95.18     &        & 96.99    & 93.42     &        & 2.48     & 2.95      &        & 0.35     & 0.49      &        \\
                                  & nnU-Net         & 97.82    & 96.18     &        & \underline{97.15}    & 94.49     &        & \textbf{1.16}     & \underline{2.27}      &        & \textbf{0.19}     & \underline{0.33}      &        \\
                                  & Swin-Unet       & 93.08    & 91.88     &        & 90.57    & 86.84     &        & 6.11     & 7.82      &        & 1.47     & 1.42      &        \\
                                  & FAS-UNet        & 95.88    & \underline{97.43}     &        & 94.92    & \underline{95.62}     &        & 1.58     & 3.84      &        & 0.25     & 0.59      &        \\
                                  & PottsMGNet      & 93.71    & 96.01     &        & 92.54    & 93.88     &        & 3.53     & 4.03      &        & 0.98     & 0.66      &        \\
                                  & \textbf{GUMP\,(Ours)}      & \textbf{99.08}    & \textbf{98.31}     &        & \textbf{98.29}    & \textbf{96.73}     &        & \underline{1.26}     & \textbf{1.87}      &        & \underline{0.20}     & \textbf{0.24}      &        \\ \hline
\end{tabular}
\end{table}

\begin{figure}[htbp]
\centering
\includegraphics[width=0.99\textwidth]{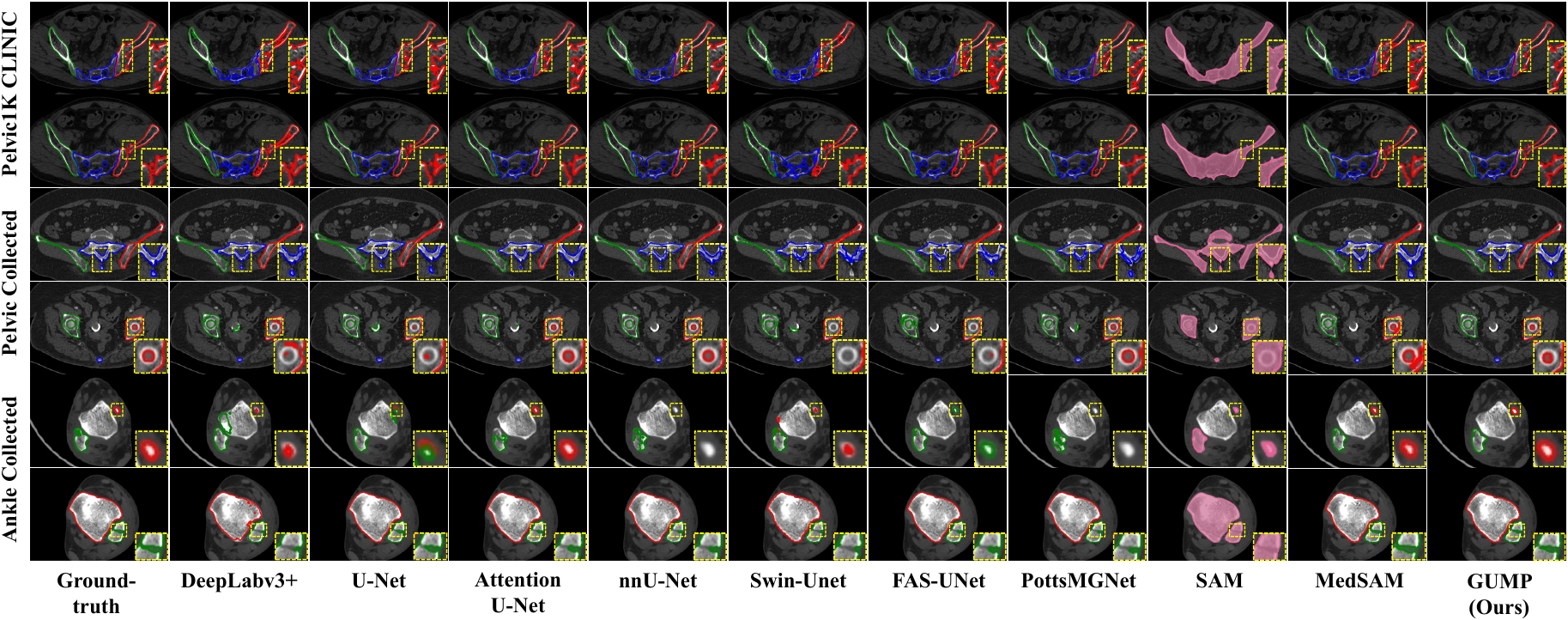}
\caption{Visual comparisons with other methods. For the dataset Pelvic1K CLINIC and Pelvic Collected, the red, green and blue contours denote the mask boundary of the left hip, right hip and sacrum respectively. For Ankle Collected, the red and green contours denote the mask boundary of the tibia and fibula respectively. Images are cropped to the concerned region for display convenience.}
\label{wang8}
\end{figure}

In the visual comparison illustrated in Fig. \ref{wang8}, the results of SAM and MedSAM are chosen the best among multiple interactions. The results of SAM are screen-captured on the official website \footnote{Available at: https://aidemos.meta.com/segment-anything}. As for MedSAM, we first try the model parameters given on the github \footnote{Available at: https://github.com/bowang-lab/MedSAM} but fail to get desirable results, which might be explained by the lack of bone datasets used in the fine-tuning. In Fig. \ref{wang8}, the results of MedSAM are obtained after 20 epochs fine-tuning on our own datasets. Comparatively, GUMP-Net can find ambiguous boundaries in bone fractures, handle complex topologies in fine structures like sacrum, and effectively reduce segmentation outliers. 

\subsection{Impact of Dataset Size}
We further investigate how dataset size impacts the segmentation performance of the compared methods by the following two experiments within the Pelvic Collected dataset.

\begin{figure}[htbp]
\centering
\includegraphics[width=0.99\textwidth]{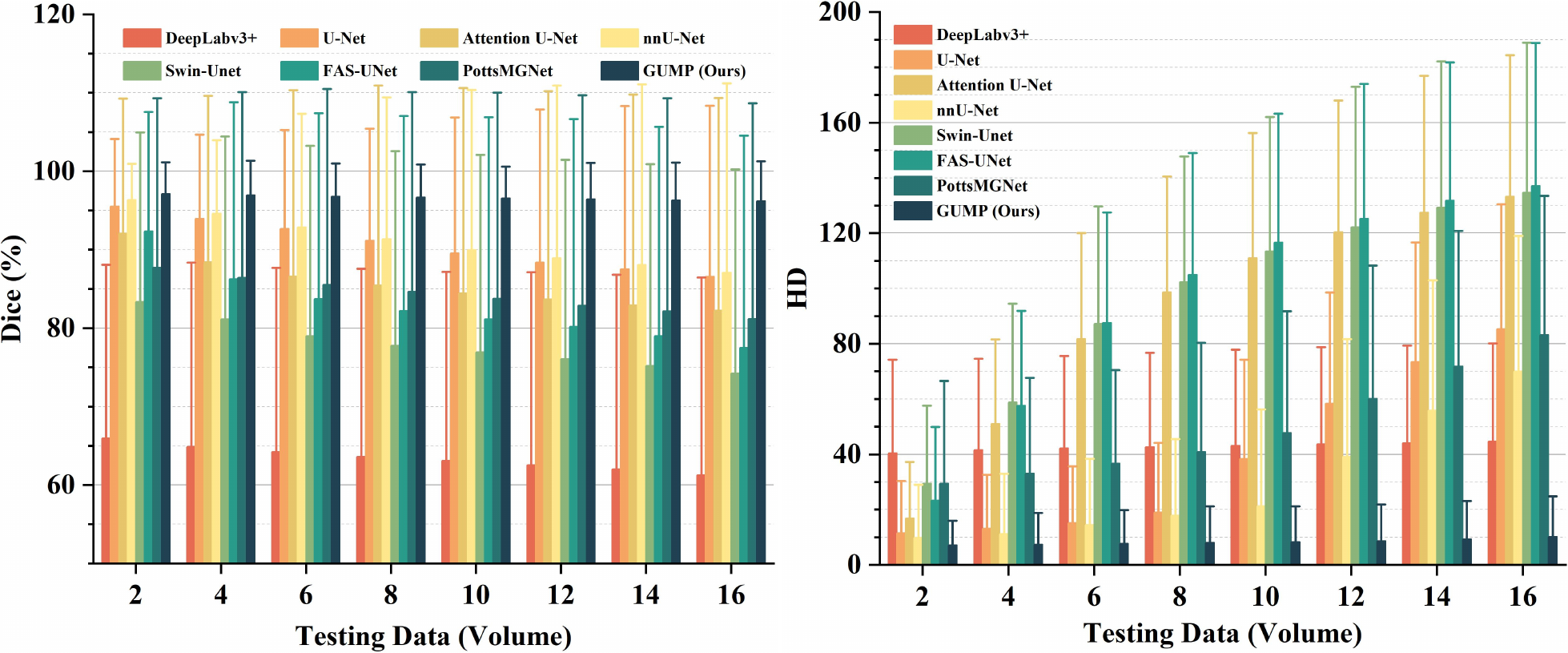}
\caption{Generalization ability of different methods with small training data.}
\label{wang6}
\end{figure}

First, we focus on the generalization ability within the dataset with small training data. To this end, we train all methods with only two volumes of training data and test on increasing volumes of testing data. Fig. \ref{wang6} shows that GUMP-Net maintains stability and consistency in Dice and HD even when the testing data becomes much larger than the training data.

\begin{figure}[htbp!]
\centering
\includegraphics[width=0.99\textwidth]{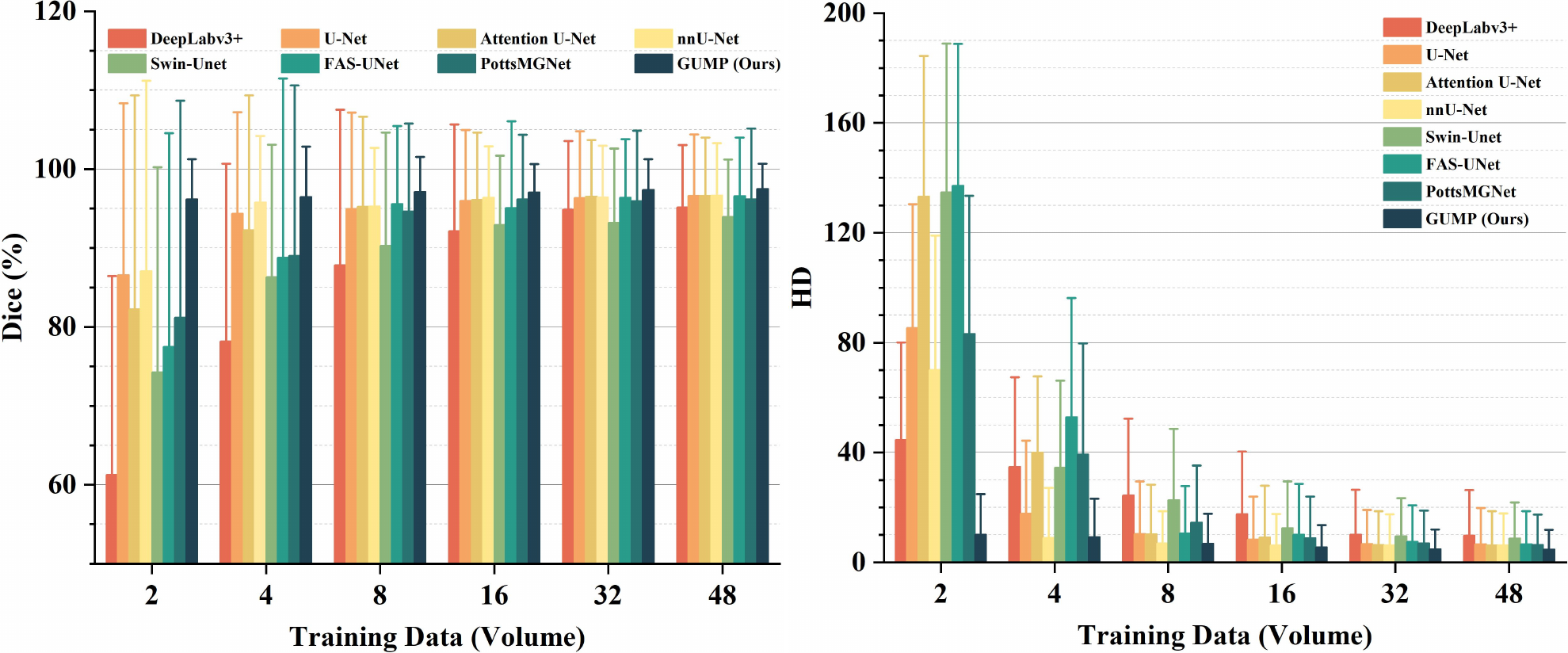}
\caption{Scaling ability of different methods with increasing training data.}
\label{wang7}
\end{figure}

Second, we consider how the performance scales with training dataset size. To do this, we train all methods with increasing volumes of training data and test on 16 volumes of testing data. Results in Fig. \ref{wang7} show that as the training data grows, all methods improve to a certain extend. However, GUMP-Net have already achieved relatively high accuracy with small training data, while other methods have just caught up with the performance of GUMP-Net until 48 volumes of training data. 

\section{Conclusion}
\label{sec:conclusion}
In this work, we proposed GUMP-Net, an interpretable model-data-driven multi-class pelvic segmentation framework that unrolls an improved GAC model into trainable CNN modules, including the object detection module (ODM), the edge detector module (EDM), and the iteration module (IM). The ODM facilitates automatic level set initialization. The EDM, based on a novel edge detector function (EDF), is designed to automatically learn the orthopedic knowledge, thereby providing an anatomy-aware EDF that can highlight bone features and suppress unconcerned regions. In the end, the IM evolves the initial level set function under the guidance of the learned EDF. Extensive experiments on pelvic and ankle CT datasets demonstrate the following advantages.

\textbf{Performance Superiority.} GUMP-Net achieves superior performance in terms of accuracy, consistency and robustness across tested datasets and anatomical regions, especially in small training data situation. This superiority benefits from the following aspects: accurate initialization provides additional spatial information and helps to narrow down the segmentation region; the learned anatomy-aware EDF helps the contours evolve without obstruction; the level set function contains both regional and global information, leading to higher sensitivity to segmentation errors. 

\textbf{Interpretability.} GUMP-Net's module design reflects the segmentation process of an improved GAC model. Each component has a clear physical meaning and makes the whole system interpretable. 

\begin{figure}[htbp]
\centering
\includegraphics[width=0.9\textwidth]{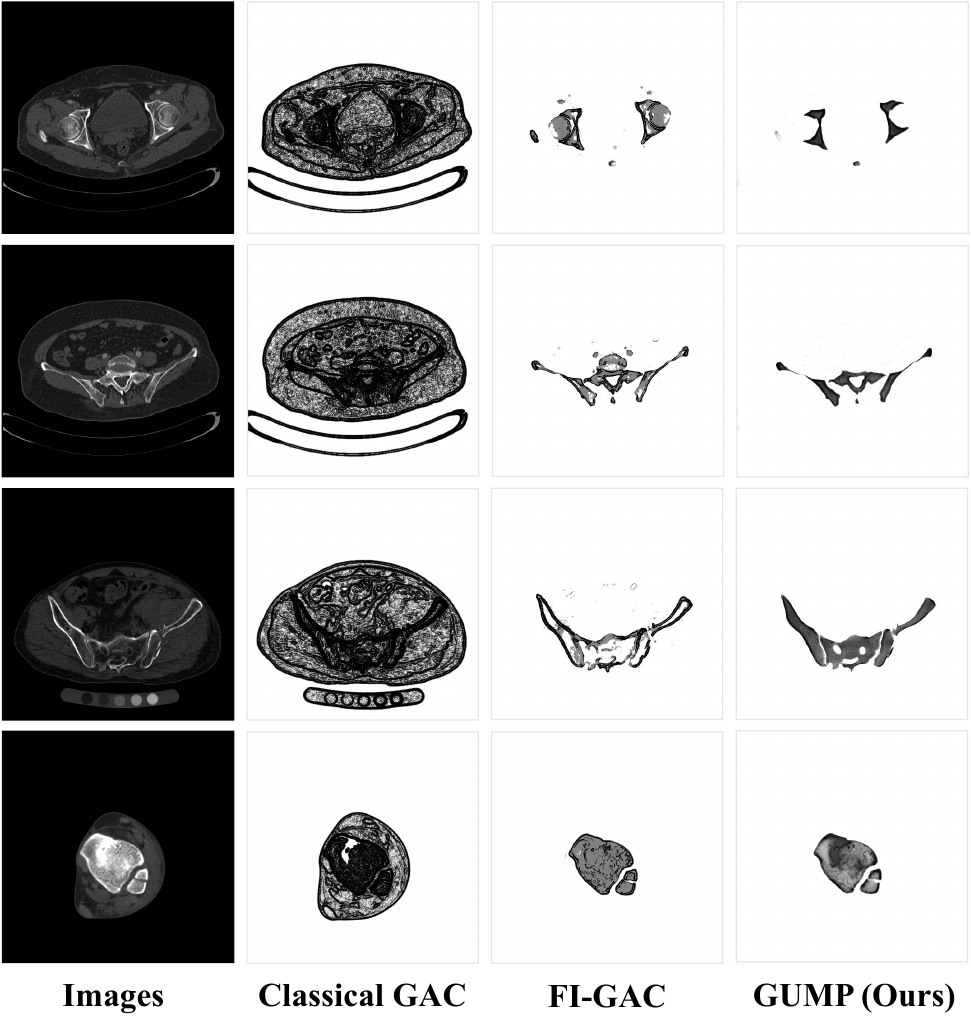}
\caption{Visual comparisons between the edge detector functions provided by the classical GAC, FI-GAC and the proposed GUMP-Net.}
\label{wang3}
\end{figure}

The proposed EDM is designed based on the hand-crafted EDF in FI-GAC, where the embedded orthopedic knowledge is learned by a neural network instead. Visual comparisons between the EDFs provided by the classical GAC, FI-GAC and the proposed EDM are illustrated in Fig. \ref{wang3}. It is observed that, while FI-GAC filters out the soft-tissue by hand-crafted parameters, the learned EDF further improves the clarity and even suppresses unconcerned bone structures by leveraging the knowledge from data. The learned EDF no longer depends on the hand-crafted parameters and can be adapted to different images. Additionally, the learned EDF can also feed back the traditional variational model positively. 

\begin{figure}[htbp]
\centering
\includegraphics[width=0.9\textwidth]{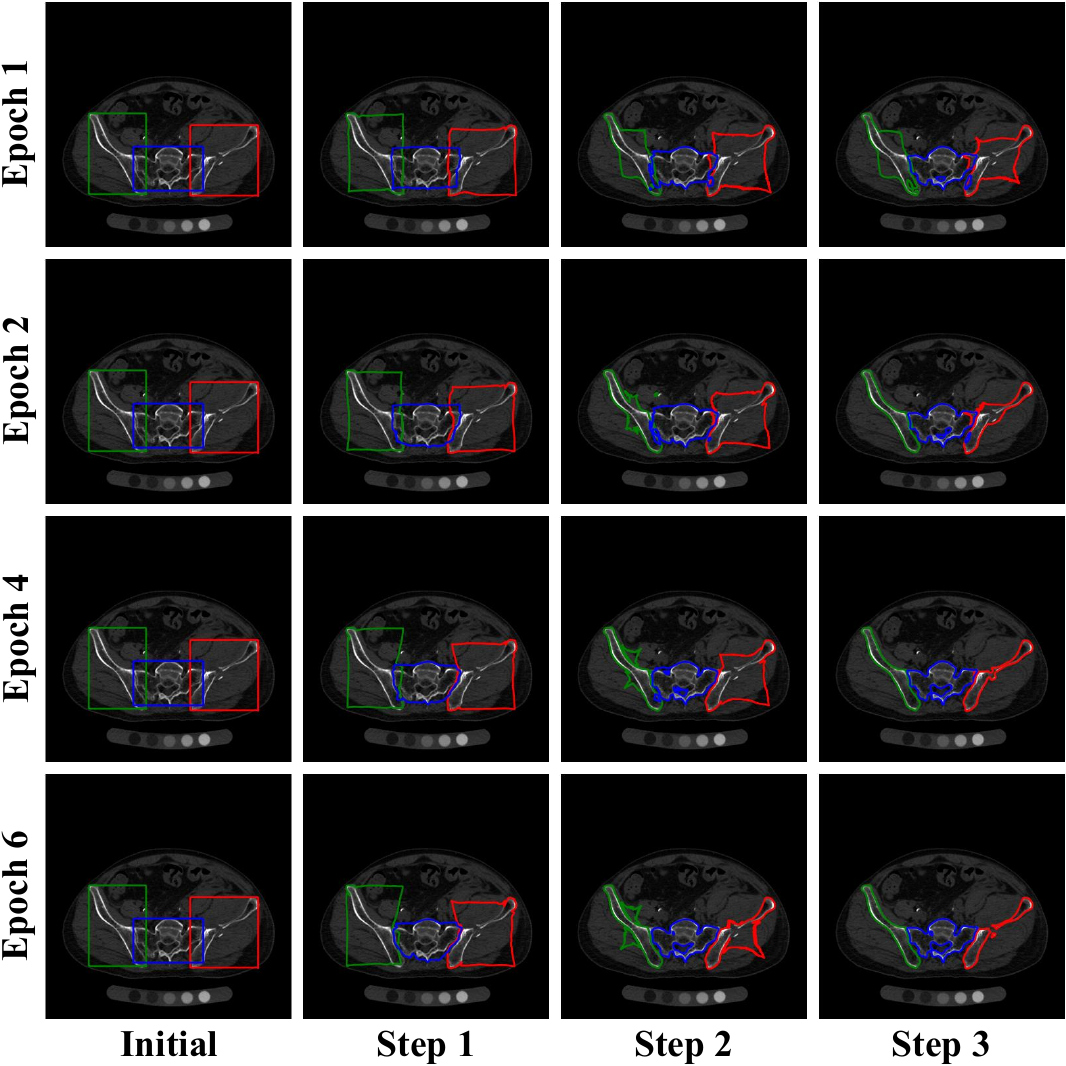}
\caption{The contour evolution across different training epochs.}
\label{wang4}
\end{figure}

The proposed IM simulates the iteration process of FI-GAC by neural networks. A visualization of the contour evolution across different training epochs is illustrated in Fig. \ref{wang4}. As both iteration step and training epoch increase, the contours tend to evolve towards the target boundaries, which means the network gradually learns how to evolve the contours to reach the target boundaries in finite steps. 

\textbf{Initialization Flexibility.} As mentioned in subsection \ref{Ablation}, GUMP-Net can be trained and tested under different bounding box settings, which can enable flexible initialization and adapt to various practical applications. Specifically, when neither pre-trained ODM nor manual prompt is available, GUMP-Net can be trained with pre-defined fixed loose bounding boxes and needs no more other initialization in the inference, thereby reducing human labors. When either pre-trained ODM or manual prompt is available, GUMP-Net can be trained with tight bounding boxes and the proposed training technique ``Bounding Box Random Shift'' enables various valid initial contours in the inference. In this situation, when the initial contours given by the ODM are not trusted by the clinicians, human interventions and corrections still remain possible. For example, if the ODM fails to detect extremely small structures, segmentation results can be obtained by initializing with manual bounding boxes.

Admittedly, there are still some limitations. First, the proposed GUMP-Net needs to compute signed distance function in the pipeline, which brings some computational burden during network training. Second, the current network has not yet been trained on fracture datasets with metal artifacts, which are common in clinical practice. Nonetheless, by decomposing the overall segmentation task into interpretable sub-tasks, our method offers the clinicians an intuitive understanding of the deep segmentation process, addressing the demand of interpretable medical artificial intelligence. Future work will explore dual-agent deep decision-making and full process intervenable neural networks.

\section*{Declaration}
The authors declare that no generative AI tools were used at any stage of the preparation of this manuscript.

\section*{Acknowledgments}
The work of YZ, LZ, HX and QC was partially supported by National Key Research $\&$ Development Program of China with grant number 2022YFC2504300. The work of LW was partially supported by National Key Research $\&$ Development Program of China with grant number  2023YFA1009300. The work of CC was partially supported by National Natural Science Foundation of China with grant numbers 12322117 and 12288201.

\bibliographystyle{plain}
\bibliography{reference}

@article{han2021fracture,
  title={Fracture reduction planning and guidance in orthopaedic trauma surgery via multi-body image registration},
  author={Han, R. and Uneri, A. and Vijayan, R. C. and Wu, P. and Vagdargi, P. and Sheth, N. and Vogt, S. and Kleinszig, G. and Osgood, G. M. and Siewerdsen, J. H.},
  journal={{Med. Image Anal.}},
  volume={68},
  pages={101917},
  year={2021},
  publisher={Elsevier}
}

@article{Liu2020,
  title={{Deep learning to segment pelvic bones: large-scale CT datasets and baseline models}},
  author={Liu, P. and Han, H. and Du, Y. and Zhu, H. and Li, Y. and Gu, F. and Xiao, H. and Li, J. and Zhao, C. and  Xiao, L. and Wu X. and Zhou, S.},
  journal={{Int. J. Comput. Assist. Radiol. Surg.}},
  year={2020},
  volume={16},
  pages={749-756},
}

@InProceedings{Liu2023,
author="Liu, Y.
and Yibulayimu, S.
and Sang, Y.
and Zhu, G.
and Wang, Y.
and Zhao, C.
and Wu, X.",
title={{Pelvic Fracture Segmentation Using a Multi-scale Distance-Weighted Neural Network}},
booktitle={{Proc. Int. Conf. Med. Image Comput. Comput.-Assist. Intervent (MICCAI)}},
year="2023",
pages="312-321",
isbn="978-3-031-43996-4"
}

@article{Toruner2024,
  author    = {Toruner, M. D. and Wang, Y. and Jiao, Z. and Bai, H.},
  title     = {{Artificial Intelligence in Radiology: Where Are We Going?}},
  journal   = {{eBioMedicine}},
  volume    = {109},
  year      = {2024},
  publisher = {Elsevier},
  issn      = {2352-3964},
}

@article{Tikhomirov2024,
  author    = {Tikhomirov, L. and Semmler, C. and McCradden, M. and Searston, R. and Ghassemi, M. and Oakden-Rayner, L.},
  title     = {{Medical Artificial Intelligence for Clinicians: The Lost Cognitive Perspective}},
  journal   = {{The Lancet Digit. Health}},
  volume    = {6},
  number    = {8},
  pages     = {589-594},
  year      = {2024},
  publisher = {Elsevier},
  issn      = {2589-7500},
}

@ARTICLE{Chan2001,
  author={Chan, T. F. and Vese, L. A.},
  journal={{IEEE Trans. Image Process.}}, 
  title={{Active Contours without Edges}}, 
  year={2001},
  volume={10},
  number={2},
  pages={266-277},
}

@INPROCEEDINGS{Caselles1995,
  author={Caselles, V. and Kimmel, R. and Sapiro, G.},
  booktitle={{Proc. IEEE/CVF Int. Conf. Comput. Vis. (ICCV)}}, 
  title={{Geodesic Active Contours}}, 
  year={1995},
  volume={},
  number={},
  pages={694-699},
}

@article{Chen2017,
  title={{A General Framework of Piecewise-polynomial Mumford--Shah Model for Image Segmentation}},
  author={Chen, C. and Leng, J. and Xu, G.},
  journal={{Int. J. Comput. Math.}},
  year={2017},
  volume={94},
  pages={1981-1997},
}

@ARTICLE{Li2008,
  author={Li, C. and Kao, C.-Y. and Gore, J. C. and Ding, Z.},
  journal={{IEEE Trans. Image Process.}}, 
  title={{Minimization of Region-Scalable Fitting Energy for Image Segmentation}}, 
  year={2008},
  volume={17},
  number={10},
  pages={1940-1949}
}

@article{Han2020,
title = {{Active contour model for inhomogenous image segmentation based on Jeffreys divergence}},
journal = {{Pattern Recogn.}},
volume = {107},
pages = {107520},
year = {2020},
issn = {0031-3203},
author = {Han, B. and Wu, Y.}
}

@INPROCEEDINGS{Jia2006,
  author={Jia, Y. and Jiang, Y.},
  booktitle={{Int. Conf. Comput. Graphics, Imaging and Visual.}}, 
  title={{Active Contour Model with Shape Constraints for Bone Fracture Detection}}, 
  year={2006},
  volume={},
  number={},
  pages={90-95}
}

@article{WANG2026113049,
title = {Fracture interactive geodesic active contours for bone segmentation},
journal = {{Pattern Recogn.}},
volume = {175},
pages = {113049},
year = {2026},
issn = {0031-3203},
author = {Wang, L. and Zhang, L. and Xu, H. and Zhao, J. and Su, X. and Li, J. and Tang, M. and Gao, W. and Chen, C.},
}

@InProceedings{Ronneberger2015,
author="Ronneberger, O.
and Fischer, P.
and Brox, T.",
title={{U-Net: Convolutional Networks for Biomedical Image Segmentation}},
booktitle={{Proc. Int. Conf. Med. Image Comput. Comput.-Assist. Intervent (MICCAI)}},
year="2015",
pages="234-241"
}

@article{Oktay2018,
  title={{Attention U-Net: Learning Where to Look for the Pancreas}},
  author={Oktay, O. and Schlemper, J. and Folgoc, L. and Lee, M. and Heinrich, M. and Misawa, K. and Mori, K. and McDonagh, S. and others},
  journal={ArXiv},
  year={2018},
  volume={abs/1804.03999}
}

@article{Isensee2021,
  author    = {F. Isensee and P. F. Jaeger and S. A. A. Kohl and J. Petersen and K. H. Maier-Hein},
  title     = {{nnU-Net: A Self-Configuring Method for Deep Learning-Based Biomedical Image Segmentation}},
  journal   = {{Nat. Methods}},
  volume    = {18},
  number    = {2},
  pages     = {203-211},
  year      = {2021},
  issn      = {1548-7105},
  publisher = {Nature Publishing Group}
}

@inproceedings{Hu2021,
  title={{Swin-Unet: Unet-like Pure Transformer for Medical Image Segmentation}},
  author={Cao, H. and Wang, Y. and Chen, J. and Jiang, D. and Zhang, X. and Tian, Q. and Wang, M.},
  booktitle={{ECCV}},
  year={2021},
  pages={205-218}
}

@ARTICLE{Zhou2020,
  author={Zhou, Z. and Siddiquee, M. and Tajbakhsh, N. and Liang, J.},
  journal={{IEEE Trans. Med. Imag.}}, 
  title={{UNet++: Redesigning Skip Connections to Exploit Multiscale Features in Image Segmentation}}, 
  year={2020},
  volume={39},
  number={6},
  pages={1856-1867}
}

@ARTICLE{Li2018,
  author={Li, X. and Chen, H. and Qi, X. and Dou, Q. and Fu, C.-W. and Heng, P.-A.},
  journal={{IEEE Trans. Med. Imag.}}, 
  title={{H-DenseUNet: Hybrid Densely Connected UNet for Liver and Tumor Segmentation From CT Volumes}}, 
  year={2018},
  volume={37},
  number={12},
  pages={2663-2674}
}

@inproceedings{Chen2018,
  title={{Encoder-Decoder with Atrous Separable Convolution for Semantic Image Segmentation}},
  author={Chen, L.-C.  and Zhu, Y. and Papandreou, G. and Schroff, F. and Adam, H.},
  booktitle={{ECCV}},
  year={2018},
  pages="833-851"
}

@inproceedings{Vaswani2017,
  title={{Attention is All you Need}},
  author={Vaswani, A. and Shazeer, N. M. and Parmar, N. and Uszkoreit, J. and Jones, L. and Gomez, A. N. and Kaiser, L. and Polosukhin, I.},
  booktitle={{NeurIPS}},
  year={2017},
  pages = "6000-6010"
}

@article{Liu2021,
  title={{Swin Transformer: Hierarchical Vision Transformer using Shifted Windows}},
  author={Liu, Z. and Lin, Y. and Cao, Y. and Hu, H. and others},
  journal={{Proc. IEEE/CVF Int. Conf. Comput. Vis. (ICCV)}},
  year={2021},
  pages={9992-10002}
}

@Inbook{HoangNganLe2020,
author="Hoang Ngan Le, T.
and Luu, K.
and Duong, C. N.
and Quach, K. G.
and Truong, T. D.
and Sadler, K.
and Savvides, M.",
title={{Active Contour Model in Deep Learning Era: A Revise and Review}},
bookTitle="Applications of Hybrid Metaheuristic Algorithms for Image Processing",
year="2020",
publisher="Springer International Publishing",
address="Cham",
pages="231-260",
isbn="978-3-030-40977-7"
}

@Inbook{Gui2023,
author="Gui, L.
and Ma, J.
and Yang, X.",
title="Variational Models and Their Combinations with Deep Learning in Medical Image Segmentation: A Survey",
booktitle="Handbook of Mathematical Models and Algorithms in Computer Vision and Imaging: Mathematical Imaging and Vision",
year="2023",
publisher="Springer International Publishing",
address="Cham",
pages="1001-1022",
isbn="978-3-030-98661-2"
}

@article{Yang2024,
  author    = {L. Yang and D. Shao and Z. Huang and M. Geng and N. Zhang and L. Chen and X. Wang and D. Liang and Z. F. Pang and Z. Hu},
  title     = {{Few-Shot Segmentation Framework for Lung Nodules via an Optimized Active Contour Model}},
  journal   = {{Med. Phys.}},
  volume    = {51},
  number    = {4},
  pages     = {2788-2805},
  year      = {2024},
  publisher = {Wiley}
}

@article{Tian2023,
year = {2023},
publisher = {IOP Publishing},
volume = {68},
number = {14},
pages = {145005},
author = {Tian, L. and Zou, L. and Yang, X.},
title = {A two-stage data-model driven pancreas segmentation strategy embedding directional information of the boundary intensity gradient and deep adaptive pointwise parameters},
journal = {{Phys. Med. Biol.}}
}

@InProceedings{Hatamizadeh19,
author="Hatamizadeh, A.
and Hoogi, A.
and Sengupta, D.
and Lu, W.
and Wilcox, B.
and Rubin, D.
and Terzopoulos, D.",
title="Deep Active Lesion Segmentation",
booktitle="Machine Learning in Medical Imaging",
year="2019",
publisher="Springer International Publishing",
address="Cham",
pages="98-105"
}

@ARTICLE{Ma2021b,
  author={Ma, J. and He, J. and Yang, X.},
  journal={{IEEE Trans. Med. Imag.}}, 
  title={{Learning Geodesic Active Contours for Embedding Object Global Information in Segmentation CNNs}}, 
  year={2021},
  volume={40},
  number={1},
  pages={93-104},
  }

@ARTICLE{Kim2020,
  author={Kim, B. and Ye, J. C.},
  journal={{IEEE Trans. Image Process.}}, 
  title={{Mumford–Shah Loss Functional for Image Segmentation With Deep Learning}}, 
  year={2020},
  volume={29},
  number={},
  pages={1856-1866}
}

@article{Liu2022,
  author    = {Liu, J. and Wang, X. and Tai, X.-C.},
  title     = {{Deep Convolutional Neural Networks with Spatial Regularization, Volume and Star-Shape Priors for Image Segmentation}},
  journal   = {{J. Math. Imaging Vis.}},
  year      = {2022},
  volume    = {64},
  number    = {6},
  pages     = {625-645},
  issn      = {1573-7683}
}

@article{Jin2024,
author = {Jin, Z. and Wang, H. and Ng, M. and Min, L.},
title = {{Regularized CNN with Geodesic Active Contour and Edge Predictor for Image Segmentation}},
journal = {{SIAM J. Imaging Sci.}},
volume = {17},
number = {4},
pages = {2392-2417},
year = {2024}
}

@ARTICLE{Monga2021,
  author={Monga, V. and Li, Y. and Eldar, Y. C.},
  journal={{IEEE Signal Process. Mag.}}, 
  title={{Algorithm Unrolling: Interpretable, Efficient Deep Learning for Signal and Image Processing}}, 
  year={2021},
  volume={38},
  number={2},
  pages={18-44},
 }

@Article{Zhang2022,
AUTHOR = {Zhu, H. and Shu, S. and Zhang, J.},
TITLE = {{FAS-UNet: A Novel FAS-Driven UNet to Learn Variational Image Segmentation}},
JOURNAL = {Mathematics},
VOLUME = {10},
YEAR = {2022},
NUMBER = {21},
ARTICLE-NUMBER = {4055},
ISSN = {2227-7390}
}

@article{Tai2024,
author = {Tai, X.-C. and Liu, H. and Chan, R.},
title = {{PottsMGNet: A Mathematical Explanation of Encoder-Decoder Based Neural Networks}},
journal = {{SIAM J. Imaging Sci.}},
volume = {17},
number = {1},
pages = {540-594},
year = {2024}
}

@InProceedings{Zhang2020,
author="Zhang, M.
and Dong, B.
and Li, Q.",
title={{Deep Active Contour Network for Medical Image Segmentation}},
booktitle={{Proc. Int. Conf. Med. Image Comput. Comput.-Assist. Intervent (MICCAI)}},
year="2020",
pages="321-331",
isbn="978-3-030-59719-1"
}

@InProceedings{Hatamizadeh2020,
author="Hatamizadeh, A.
and Sengupta, D.
and Terzopoulos, D.",
title={{End-to-End Trainable Deep Active Contour Models for Automated Image Segmentation: Delineating Buildings in Aerial Imagery}},
booktitle={{ECCV}},
year="2020",
pages="730-746",
isbn="978-3-030-58610-2"
}

@INPROCEEDINGS{Redmon2016,
  author={Redmon, J. and Divvala, S. and Girshick, R. and Farhadi, A.},
  booktitle={{Proc. IEEE/CVF Conf. Comput. Vis. Pattern Recognit. (CVPR)}}, 
  title={{You Only Look Once: Unified, Real-Time Object Detection}}, 
  year={2016},
  volume={},
  number={},
  pages={779-788},
  }

@article{Peng2020DeepSF,
  title={{Deep Snake for Real-Time Instance Segmentation}},
  author={Peng, S. and Jiang, W. and Pi, H. and Bao, H. and Zhou, X.},
  journal={{Proc. IEEE/CVF Conf. Comput. Vis. Pattern Recognit. (CVPR)}},
  year={2020},
  pages={8530-8539},
}

@article{Yang2023BoxSnakePI,
  title={{BoxSnake: Polygonal Instance Segmentation with Box Supervision}},
  author={Yang, R. and Song, L. and Ge, Y. and Li, X.},
  journal={{Proc. IEEE/CVF Int. Conf. Comput. Vis. (ICCV)}},
  year={2023},
  pages={766-776},
}

@article{SAM,
  title={{Segment Anything}},
  author={Kirillov, A. and Mintun, E. and Ravi, N. and Mao,H. and Rolland, C. and Gustafson, L. and Xiao, T. and Whitehead, S. and Berg, A. C. and Lo, W.-Y. and Doll{\'a}r, P. and Girshick, R. B.},
  journal={{Proc. IEEE/CVF Int. Conf. Comput. Vis. (ICCV)}},
  year={2023},
  pages={3992-4003}
}

@article{MEDSAM,
  title={{Segment Anything in Medical Images}},
  author={Ma, J. and He, Y. and Li, F. and Han, L.-J. and You, C. and Wang, B.},
  journal={Nat. Commun.},
  year={2023},
  volume={15}
}

@article{SAB,
title = {{SegmentAnyBone: A universal model that segments any bone at any location on MRI}},
journal = {{Med. Image Anal.}},
volume = {101},
pages = {103469},
year = {2025},
issn = {1361-8415},
author = {Gu, H. and Colglazier, R. and Dong, H. and Zhang, J. and Chen, Y. and Yildiz, Z. and others},
}

@article{MIN201969,
title = {A multi-scale level set method based on local features for segmentation of images with intensity inhomogeneity},
journal = {{Pattern Recogn.}},
volume = {91},
pages = {69-85},
year = {2019},
author = {Min, H. and Xia, L. and Han, J. and Wang, X. and others}
}

@article{ZHANG2026112697,
title = {{MTD-Net: A robust multi-task discriminative network for choroidal neovascularization segmentation}},
journal = {{Pattern Recogn.}},
volume = {172},
pages = {112697},
year = {2026},
issn = {0031-3203},
author = {Zhang, D. and Liu, M. and Chen, T. and Li, H. and Ying, J. and Chen, D. and Li, B. and Yi, Q. and Zhang, J.}
}

@article{HAN2023109076,
title = {{ML-DSVM+: A meta-learning based deep SVM+ for computer-aided diagnosis}},
journal = {{Pattern Recogn.}},
volume = {134},
pages = {109076},
year = {2023},
issn = {0031-3203},
author = {Han, X. and Wang, J. and Ying, S. and Shi, J. and Shen, D.}
}

@article{LIU2026112384,
title = {{A variable gaussian kernel scale active contour model based on Jeffreys divergence for ICT image segmentation}},
journal = {{Pattern Recogn.}},
volume = {172},
pages = {112384},
year = {2026},
issn = {0031-3203},
author = {Liu, Z. and Li, Q. and Wang, J. and Deng, T. and Zhou, R. and Cai, Y. and Liu, F.}
}

\end{document}